\documentclass[letterpaper]{article} 
\usepackage[submission]{aaai25}  
\usepackage{times}  
\usepackage{helvet}  
\usepackage{courier}  
\usepackage[hyphens]{url}  
\usepackage{graphicx} 
\usepackage{natbib}  
\usepackage{caption} 
\usepackage{algorithm}
\usepackage{algorithmicx}
\usepackage{algpseudocode}
\usepackage{enumitem}
\usepackage{multirow}

\usepackage{newfloat}
\usepackage{listings}
\DeclareCaptionStyle{ruled}{labelfont=normalfont,labelsep=colon,strut=off} 
\floatstyle{ruled}
\newfloat{listing}{tb}{lst}{}
\floatname{listing}{Listing}
\usepackage{color}
\usepackage{amsmath}
\usepackage{amsfonts}

\usepackage{amssymb}
\newlist{checkboxlist}{itemize}{1}
\setlist[checkboxlist,1]{label=$\square$}
\usepackage{pifont}

\title{{CP-Guard}: Malicious Agent Detection and Defense in Collaborative Bird's Eye View Perception}
\author{
    Senkang Hu\textsuperscript{\rm 1}\equalcontrib, 
    Yihang Tao\textsuperscript{\rm 1}\equalcontrib,
    Guowen Xu\textsuperscript{\rm 2},
    Yiqin Deng\textsuperscript{\rm 1}\thanks{Corresponding author.}, 
    Xianhao Chen\textsuperscript{\rm 3},    \\
    Yuguang Fang\textsuperscript{\rm 1}, and Sam Kwong\textsuperscript{\rm 4}
}
\affiliations{
    \textsuperscript{\rm 1}Department of Computer Science, City University of Hong Kong\\
    \textsuperscript{\rm 2}School of Computer Science and Engineering, University of Electronic Science and Technology of China\\
    \textsuperscript{\rm 3}Department of Electrical and Electronic Engineering, The University of Hong Kong\\
    
    \textsuperscript{\rm 4}Department of Computing and Decision Sciences, Lingnan University\\
    { 
    {\{senkang.forest, yihang.tommy\}@my.cityu.edu.hk, guowen.xu@uestc.edu.cn}\\ 
    {xchen@eee.hku.hk, samkwong@ln.edu.hk, \{yiqideng, my.Fang\}@cityu.edu.hk}
    }

}

\usepackage{bibentry}

\begin{document}

\maketitle

\begin{abstract}

Collaborative Perception (CP) has shown a promising technique for autonomous driving, where multiple connected and autonomous vehicles (CAVs) share their perception information to enhance the overall perception performance and expand the perception range. However, in CP, ego CAV needs to receive messages from its collaborators, which makes it easy to be attacked by malicious agents. For example, a malicious agent can send harmful information to the ego CAV to mislead it. To address this critical issue, we propose a novel method, \textbf{CP-Guard}, a tailored defense mechanism for CP that can be deployed by each agent to accurately detect and eliminate malicious agents in its collaboration network. Our key idea is to enable CP to reach a consensus rather than a conflict against the ego CAV's perception results. Based on this idea, we first develop a probability-agnostic sample consensus (PASAC) method to effectively sample a subset of the collaborators and verify the consensus without prior probabilities of malicious agents. 
Furthermore, we define a collaborative consistency loss (CCLoss) to capture the discrepancy between the ego CAV and its collaborators, which is used as a verification criterion for consensus.
Finally, we conduct extensive experiments in collaborative bird's eye view (BEV) tasks and our results demonstrate the effectiveness of our CP-Guard. Code is available at \texttt{https://github.com/CP-Security/CP-Guard}.

\end{abstract}


\section{Introduction}
\label{sec:intro}

\begin{figure}[t]
    \centering
    \includegraphics[width=0.9\linewidth]{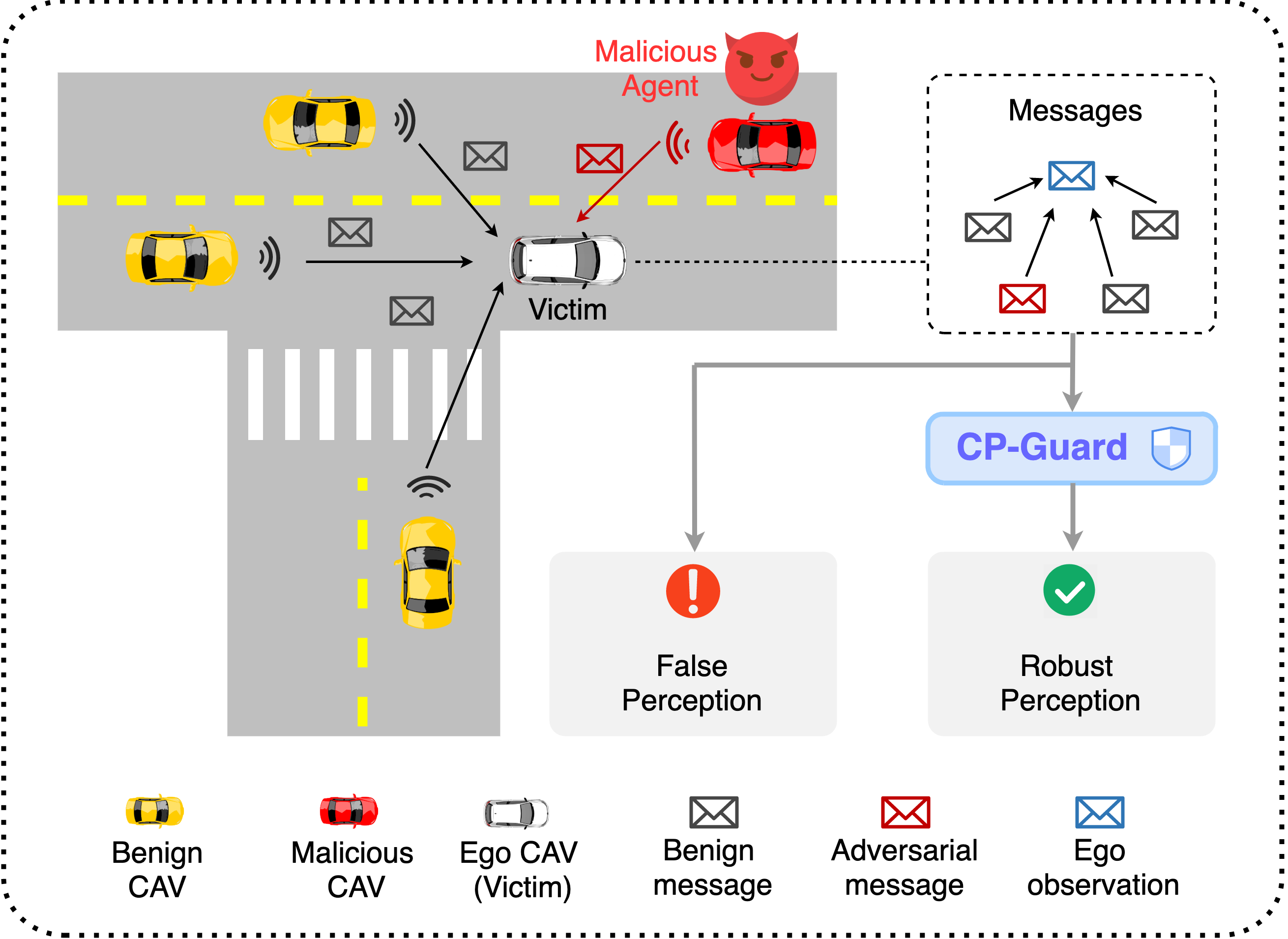}
    \caption{\textbf{Illustration of the threats of malicious agent in collaborative perception and our defense framework, CP-Guard}. When there is no defense, malicious CAVs could easily send intricately crafted adversarial messages to the ego CAV, consequently misleading the CP system and resulting in false perception outputs. To counter this vulnerability, we propose CP-Guard,  a tailored defense mechanism for CP that can effectively detect and neutralize malicious agents, thereby ensuring robust perception outcomes.}
    \label{fig:security_threats}
    \vspace{-3mm} 
\end{figure}

Recently, multi-agent collaborative perception has attracted great attention from both academia and industries since it can overcome the limitation of single-agent perception such as occlusion and limitation of sensing range \cite{hanCollaborativePerceptionAutonomous2023,huCollaborativePerceptionConnected2024}. Because the collaborative CAVs send complementary information (e.g., raw sensor data, intermediate features, and perception results) to ego CAV, the ego CAV can leverage this complementary information to extend its perception range and tackle the blind spot problem in its view, which is crucial for the safety of the autonomous driving systems. The operational flow of CP is as follows. Each CAV independently encodes local sensor inputs into intermediate feature maps. Then, these CAVs share their feature maps with their ego CAV by vehicle-to-vehicle (V2V) communication. Finally, the ego CAV fuses the received feature maps with its own feature maps and decodes them to acquire the final perception results.
 
However, compared with single-agent perception, multi-agent CP is more vulnerable to security threats and easy to attack, since it incorporates the information from multiple agents which makes the attack surface larger. An attack could be executed by a man-in-the-middle who alters the feature maps sent to the victim agent, or by a malicious agent that directly transmits manipulated feature maps to the victim agent. For example, Tu \textit{et al.} \cite{tuAdversarialAttacksMultiAgent2021} generated adversarial perturbations on the feature maps and attacked the ego CAV, resulting in the wrong perception results. Additionally, as the encoded feature maps are not visually interpretable by humans, moderate modifications to these maps will go unnoticed, rendering the attack quite stealthy.

This issue raises significant risks to CP if the ego CAV cannot accurately detect and eliminate the malicious agents in its collaboration network and the perception results are corrupted, which may result to catastrophic consequences. For example, the ego CAV may misclassify the traffic light status or fail to detect the front objects, leading to serious traffic accidents or even loss of life. Therefore, it is essential to develop a defense mechanism for CP that is robust to attack from malicious agents and can remove the malicious agents in its collaboration network.

In order to address this issue, several works have explored this problem. For example, Li \textit{et al.} \cite{liUsAdversariallyRobust2023} proposed robust collaborative sampling consensus (ROBOSAC) to randomly sample a subset of the collaborators and verify the consensus, but it requires the prior probabilities of malicious agents, which are usually unknown in practice. In addition, Zhao \textit{et al.} \cite{zhaoMaliciousAgentDetection2023} and Zhang \textit{et al.} \cite{zhangDataFabricationCollaborative2023} also developed defense 
methods against malicious agents, while these methods need to check the collaborators one by one, which is inefficient and computation-consuming.
Moreover, other works \cite{tuAdversarialAttacksMultiAgent2021,raghunathanUnderstandingMitigatingTradeoff2020,zhangAdversarialExamplesOpportunities2020} use adversarial training to enhance the robustness of the model. However, adversarial training introduces additional overhead during training and lacks generalization for unseen attacks. Additionally, it may result in a reduction in accuracy and is non-trivial to achieve computationally efficient and generalizable adversarial defense in CP.

In order to fill in the research gap and overcome the aforementioned limitations, we design a novel defense mechanism, CP-Guard. It can be deployed by each agent to accurately detect and eliminate malicious agents in the local collaboration network. The key idea is to enable CP to achieve a consensus rather than a conflict against the ego CAV's perception results.
Following this idea, we first design a \textit{probability-agnostic sample consensus} (PASAC) method to effectively sample a subset of the collaborators and verify the consensus without prior probabilities of malicious agents. In addition, the consensus is verified by our carefully designed \textit{collaborative consistency loss} (CCLoss), which is used to calculate the discrepancy between the ego CAV and its collaborators. 
If a collaborator's collaborative consistency loss exceeds a predefined threshold, the collaborator is considered a benign agent, otherwise, it is considered a malicious agent. The main contributions of this paper are summarized as follows. 
\begin{itemize}
    \item We analyze the vulnerabilities of CP against malicious agents and develop a novel framework for robust collaborative BEV perception, CP-Guard, which can defend against attacks and eliminate malicious agents from the local collaboration network.
    \item We establish a probability-agnostic sample consensus (PASAC) method to effectively sample a subset of the collaborators and verify the consensus without prior probabilities of malicious agents. In addition, we design a collaborative consistency loss (CCLoss) as a verification criterion for consensus, which can calculate the discrepancy between the ego CAV and its collaborators.
    \item We conduct extensive experiments on collaborative BEV tasks and the results demonstrate the effectiveness of our CP-Guard and its generalization to different attacks. 
\end{itemize}

\section{Background and Related Work}
\label{sec:background_related_work}


\subsection{Collaborative Perception}
Collaborative perception has been investigated as a means to mitigate the limitations inherent to the field-of-view (FoV) in single-agent perception systems, enhancing the accuracy, robustness, and resilience of these systems \cite{fang2024pacp, fang2024prioritizedinformationbottlenecktheoretic, fang2024racprealtimeadaptivecollaborative,qu2024mobile,qu2024trimcaching,qu2024trimcachingICDCS,linEfficientParallelSplit2024,linAdaptSFLAdaptiveSplit2024, lin2024hierarchicalsplitfederatedlearning, 10716798,linSplitLearning6G2023,linPushingLargeLanguage2023}. In this collaborative context, agents may opt for one of three predominant data fusion strategies: (1) early-stage raw data fusion, (2) intermediate-stage feature fusion, and (3) late-stage output fusion. Early-stage fusion, while increasing the data communication load, typically yields more precise collaboration outcomes. In contrast, late-stage fusion consumes less bandwidth but introduces greater uncertainty to the results. Intermediate-stage fusion, favored in much of the current literature, strikes an optimal balance between communication overhead and perceptual accuracy.
Research aimed at enhancing collaborative perception performance is multifaceted, addressing aspects such as communication overhead \cite{Su_Chen_Bai_Lin_Li_Qu_Zhou_2024, 9451536, 9312959}, robustness \cite{10160546,huAgentsCoDriverLargeLanguage2024,hu2024agentscomergelargelanguagemodel,tao2024directcpdirectedcollaborativeperception,huAdaptiveCommunicationsCollaborative2023,fang2024pibprioritizedinformationbottleneck,fang2024ic3mincarmultimodalmultiobject}, system heterogeneity \cite{lu2024an}, and domain generalization \cite{10779389}. Among these, robustness has emerged as a particularly critical focus within the field of collaborative perception.
Despite extensive studies on system intrinsic robustness, addressing challenges such as communication disruptions \cite{10457955}, pose noise correction \cite{10160546}, and communication latency \cite{10.1007/978-3-031-19824-3_19}, most existing research works have not accounted for the presence of malicious attackers within the collaborative framework. Only a selected few studies examine the implications of robustness in scenarios compromised by malicious nodes, highlighting a significant gap in current research methodologies.

\subsection{Adversarial Perception}
Adversarial attacks targeting at single-vehicle perception systems predominantly employ techniques such as GPS spoofing \cite{Li_2021_ICCV}, LiDAR spoofing \cite{279980}, and the deployment of physically realizable adversarial objects \cite{Tu_2020_CVPR, ni2023uncovering,ni2023eavesdropping,ni2023exploiting}. In the context of multi-vehicle collaborative perception, the nature of adversarial strategies can vary significantly depending on the stage of collaboration.
For early-stage collaborative perception, Zhang \textit{et al.} \cite{zhangDataFabricationCollaborative2023} have developed sophisticated attacks involving object spoofing and removal. These attacks exploit vulnerabilities by simulating the presence or absence of objects and reconstructing LiDAR point clouds using advanced ray-casting techniques. In contrast, late-stage collaboration typically involves the sharing of object locations \cite{9120490}, which provides adversaries with opportunities to manipulate these shared locations easily.
Intermediate-stage attacks are particularly nuanced, often requiring that an attacker possesses white-box access to the perception models. This knowledge enables more precise manipulations of the system, though such systems are generally resistant to simplistic black-box strategies, such as ray-casting attacks, due to the protective effect of benign feature maps which significantly reduce the efficacy of such attacks.
Tu \textit{et al.} \cite{tuAdversarialAttacksMultiAgent2021} were among the pioneers in articulating an untargeted adversarial attack aimed at maximizing the generation of inaccurate detection bounding boxes by manipulating feature maps in intermediate-fusion systems. Building on this foundational work, Zhang \textit{et al.}\cite{zhangDataFabricationCollaborative2023} have advanced the methodology by integrating perturbation initialization and feature map masking techniques to facilitate realistic, real-time targeted attacks.
Our work is dedicated to exploring and mitigating adversarial threats specifically under intermediate-level collaborative perception framework, aiming to enhance system resilience against sophisticated attacks.

\subsection{Defensive Perception}

To fortify intermediate-level collaborative perception systems against adversarial attacks, Li \textit{et al.} \cite{liUsAdversariallyRobust2023} proposed the Robust Collaborative Sampling Consensus (ROBOSAC) method. This approach entails a random selection of a subset of collaborators for consensus verification. Despite its potential, the efficacy of ROBOSAC hinges on the availability of prior probabilities of malicious intent among agents, which are often unknown in real-world scenarios.
Moreover, Zhao \textit{et al.} \cite{zhaoMaliciousAgentDetection2023} and Zhang \textit{et al.} \cite{zhangDataFabricationCollaborative2023} have formulated defensive strategies targeting at identifying malicious agents. These techniques, however, involve scrutinizing each collaborator individually, rendering them both computationally intensive and inefficient.
Adversarial training has also been explored as a mechanism to bolster system robustness, as demonstrated in the studies by Tu \textit{et al.} \cite{tuAdversarialAttacksMultiAgent2021}, Raghunathan \textit{et al.} \cite{raghunathanUnderstandingMitigatingTradeoff2020}, and Zhang \textit{et al.} \cite{zhangAdversarialExamplesOpportunities2020}. While this approach enhances system security, it substantially increases the computational load during training and may not effectively generalize to novel, unseen attacks \cite{ni2023recovering, ni2023xporter, yuan2024itpatch}. Additionally, adversarial training often results in diminished model accuracy and poses significant challenges in developing a computationally efficient and scalable adversarial defense that can be broadly applied across collaborative perception platforms.
In contrast, our research introduces a methodology that can be autonomously implemented by each agent to accurately detect and neutralize malicious entities within the local collaborative network, aiming to enhance both the efficiency and effectiveness of defense mechanisms in collaborative perception systems.
\section{Problem Setup}

\subsection{Collaborative BEV Segmentation}

BEV segmentation is essential for autonomous driving since it enables multi-modal sensor data (e.g. LiDAR point cloud, multi-view camera images) to be transformed into a unified BEV space for information aggregation or fusion. This approach offers significant advantages in accurately maintaining the spatial and temporal locations and scales of road elements. In this paper, we focus on the LiDAR-based collaborative BEV segmentation task with an intermediate feature fusion paradigm. 

Specifically, consider a set of $N + 1$ CAVs, including the ego CAV, each CAV is installed with a feature encoder $f_\text{E}$ and a feature aggregator $f_\text{A}$ as well as a BEV segmentation decoder $f_\text{D}$. 
For the $i$-th CAV, the input is a set of voxelized LiDAR point cloud  $\mathbf{X}_i\in \mathbb{R}^{W\times H \times n}$, where $W$ and $H$ are the width and height of the voxelized point cloud, respectively, and $n$ is the dimension of the voxelized point cloud. The collaborative BEV segmentation pipeline can be described as follows.

\begin{enumerate}
    \item Firstly, the feature encoder $f_\text{E}$ is used to extract the intermediate feature maps $\mathbf{F}_i = f_\text{E}(\mathbf{X}_i) \in \mathbb{R}^{\frac{W}{K}\times \frac{H}{K} \times C}$, where $K$ indicates the down-sample rate and $C$ is the number of channels. 
    \item Then, the other CAVs transmit their intermediate feature maps $\mathbf{F}_i$ to the ego CAV. The ego CAV leverages the aggregator $f_\text{A}$ to fuse the feature maps from all CAVs including its own feature map, which can be formulated as $\mathbf{F} = f_\text{A}(\mathbf{F}_0, \mathbf{F}_1, \mathbf{F}_2, \cdots, \mathbf{F}_{N}) $, where $\mathbf{F}_0$ is the ego CAV's feature map.
    \item Finally, the ego CAV decodes the aggregated feature map $\mathbf{F}$ into the final BEV segmentation map $\mathbf{Y} = f_\text{D}(\mathbf{F})$.
\end{enumerate}
During training, given the ground truth BEV segmentation map $\mathbf{Y}^*$, the loss function is defined as $ \mathcal{L}_\text{seg}(\mathbf{Y}, \mathbf{Y}^*)$. The goal is to minimize the loss function $\mathcal{L}_\text{seg}$ by optimizing the parameters of the feature encoder $f_\text{E}$, feature aggregator $f_\text{A}$, and BEV segmentation decoder $f_\text{D}$.

\begin{algorithm}[t]
    \caption{PASAC}
    \label{alg:pasac}
    \textbf{Input}: 
    \begin{itemize}[noitemsep]
        \item $\{\mathbf{F}_1, \mathbf{F}_2, \cdots, \mathbf{F}_{N}\}$, intermediate feature maps  from collaborators. $\mathbf{F}_0$, the intermediate feature of ego CAV.
        \item $f_\text{A}$, $f_\text{D}$, the aggregator and BEV segmentation decoder.
        \item $N_\text{upper}$, maximum number of selected collaborators for ego CAV, and $N_\text{upper}\leq N$.
        \item $\varepsilon$, the threshold of $\mathcal{L}_\text{CCLoss}$.
    \end{itemize}
    \textbf{Output}: $\{\mathbf{B}_i\}$, the set of benign collaborators
    \begin{algorithmic}[1] 
    \State $\mathbf{Y}_0 = f_\text{D}(\mathbf{F}_0)$.
    \Procedure{\texttt{PASAC}}{$\{\mathbf{F}_i\}_{i=1,\ldots,N}$} 
        \If{\texttt{len}$(\{\mathbf{B}_i\})\geq N_\text{upper}$}
        \State \Return $\{\mathbf{B}_i\}$
        \EndIf
        \If{\texttt{len}$(\{\mathbf{F}_i\})=1$} 
        \State $\mathbf{Y}_k = f_\text{D}(f_\text{A}(\mathbf{F}_0, \mathbf{F}_k))$.
            \If{ $\mathcal{L}_\text{CCLoss}(\mathbf{Y}_0, \mathbf{Y}_k)\leq\varepsilon$}
            \State $\{\mathbf{B}_i\} \leftarrow \{\mathbf{B}_i\}\cup\mathbf{F}_k$ 
            \EndIf
            \State \Return $\{\mathbf{B}_i\}$
        \EndIf
        \State  $\{\mathbf{F}_i\}_{i=1,\ldots,{N}}\rightarrow\{\mathbf{F}_i\}_{i=1,\ldots,\frac{N}{2}}, \{\mathbf{F}_i\}_{i=\frac{N}{2},\ldots,N}$
        \State $\mathbf{Y}_\text{G1} = f_\text{D}(f_\text{A}(\mathbf{F}_0, \{\mathbf{F}_i\}_{i=1,\ldots,\frac{N}{2}}))$ 
        \State $\mathbf{Y}_\text{G2} = f_\text{D}(f_\text{A}(\mathbf{F}_0, \{\mathbf{F}_i\}_{i=\frac{N}{2},\ldots,N}))$
        \If{$\mathcal{L}_\text{CCLoss}(\mathbf{Y}_0, \mathbf{Y}_\text{G1})\leq\varepsilon$}
        \State $\{\mathbf{B}_i\}_\text{sublist}$ = \texttt{PASAC}($\{\mathbf{F}_i\}_{i=1,\ldots,\frac{N}{2}}$)
        \State $\{\mathbf{B}_i\}\leftarrow \{\mathbf{B}_i\}\cup \{\mathbf{B}_i\}_\text{sublist}$
        \Else 
        \State $\{\mathbf{B}_i\}\leftarrow \{\mathbf{B}_i\}\cup \{\mathbf{F}_i\}_{i=1,\ldots,\frac{N}{2}}$ 
        
        \EndIf 
        \If{$\mathcal{L}_\text{CCLoss}(\mathbf{Y}_0, \mathbf{Y}_\text{G2})\leq\varepsilon$} 
        \State $\{\mathbf{B}_i\}_\text{sublist}$ = \texttt{PASAC}($\{\mathbf{F}_i\}_{i=\frac{N}{2},\ldots,N}$)
        \State $\{\mathbf{B}_i\}\leftarrow \{\mathbf{B}_i\}\cup \{\mathbf{B}_i\}_\text{sublist}$
        \Else
        \State  $\{\mathbf{B}_i\}\leftarrow\{\mathbf{B}_i\}\cup\{\mathbf{F}_i\}_{i=\frac{N}{2},\ldots,N}$ 
        
        \EndIf
        \State \Return $\{\mathbf{B}_i\}$
    \EndProcedure
    \end{algorithmic}
\end{algorithm}

\subsection{Adversarial Threat Model in CP}

In order to defend against malicious agents in CP, we first need to figure out the attack scenarios and the attacker's abilities. Specifically, we consider an attacker to have full access to malicious CAVs. In addition, since the BEV segmentation model is deployed on each CAV, the attacker has full access to the model architecture, parameters, and intermediate feature maps, enabling the attacker to launch a white-box attack. Based on this, the attacker aims to manipulate the intermediate feature maps by adding adversarial perturbations to maximize the ego CAV's BEV segmentation loss. Then, these adversarial messages are transmitted to the ego CAV to fool its perception fusion. The attacker's goal can be formulated as follows.
\begin{equation}
    \begin{aligned}
        &\max_{\|\delta\|\leq \Delta} \mathcal{L}_\text{seg}(\mathbf{Y}^\delta, \mathbf{Y}^*), \\
        &\text{\ \ s.t.}\quad \mathbf{Y}^\delta = f_\text{D}\left(f_\text{A}(\mathbf{F}_0, \mathbf{F}_1, \mathbf{F}_m +\delta, \cdots, \mathbf{F}_{N})\right),
    \end{aligned}
\end{equation}
where $\mathbf{Y}^\delta$ is the adversarial collaborative BEV segmentation map, $\mathbf{F}_m$ is the malicious agent's feature map, and $\delta$ is the optimization perturbation which is constrained by $\|\delta\|\leq \Delta$ to ensure its stealth to avoid being detected. Its size is the same as the size of intermediate feature map $\mathbf{F}_m$, which is $\mathbb{R}^{\frac{W}{K}\times \frac{H}{K} \times C}$.

\section{Method}
\label{sec:method}

In this section, we present our CP-Guard in detail. It consists of two main components: (1) \textit{Probability-Agnostic Sample Consensus} (PASAC) and (2) \textit{Collaborative Consistency Loss Verification} (CCLoss). PASAC is designed to effectively sample a subset of collaborators for consensus verification without relying on prior probabilities of malicious intent. CCLoss is proposed to verify the consensus between the ego CAV and the collaborative CAVs. These two components work collaboratively to detect and neutralize malicious agents in the collaborative perception network. We elaborate on these two components in the following subsections.

\subsection{Probability-Agnostic Sample Consensus}
\label{subsec:probability_agnostic_sample_consensus}

To achieve the consensus of collaborators, the most straightforward method is to check the collaborators one by one. However, this method is time-consuming and computation-intensive, especially when the number of collaborators is large. A better method is to randomly sample a subset of collaborators for consensus verification at each time, such as ROBOSAC \cite{liUsAdversariallyRobust2023}. However, ROBOSAC requires the prior probabilities of malicious intent among agents, which are often unknown in real-world scenarios. To fill in this research gap, we propose PASAC.

More specifically, given a set of $N$ collaborators, the ego CAV will generate the collaborative BEV segmentation map $\mathbf{Y}$ based on its observation and the received messages for feature fusion from the $N$ collaborators.
Firstly, the ego CAV generates its BEV segmentation map $\mathbf{Y}_0$ based on its own observation. Then, it randomly split the collaborators into two groups of equal size. 
After receiving all the messages, the ego CAV fuses the features and generates the BEV segmentation map $\mathbf{Y}_\text{G1}$ based on the messages $\{\mathbf{F}_i\}_{i=1,\ldots,\frac{N}{2}}$ from the first group. Similarly, it generates the BEV segmentation map $\mathbf{Y}_\text{G2}$ based on the messages $\{\mathbf{F}_i\}_{i=\frac{N}{2},\ldots, N}$ from the second group.

Then, the ego CAV verifies the consensus and checks if there is any malicious CAV in the two groups. The consensus is verified by CCLoss to be introduced in the next subsection. Specifically, the CCLoss is calculated between the ego CAV and each group, that is, $\mathcal{L}_\text{CCLoss}(\mathbf{Y}_0, \mathbf{Y}_\text{G1})$ and $\mathcal{L}_\text{CCLoss}(\mathbf{Y}_0, \mathbf{Y}_\text{G2})$. If the CCLoss exceeds a predefined threshold, the group is considered benign, otherwise, it is considered to contain malicious CAVs. 

Suppose the first group is benign and the second group is verified to have malicious CAVs, all CAVs in the first group are marked as benign and incorporated in the following collaboration. For the second group, the ego CAV continues to split the second group into two subgroups and repeats the consensus verification process. This process will continue until finding all the malicious CAVs or obtain enough benign CAVs. The detailed procedures of PASAC can be summarized as follows.
\begin{enumerate}[noitemsep]
    \item Generate the BEV segmentation map $\mathbf{Y}_0$ based on the ego CAV's observation.
    \item Split the collaborators into two groups.
    \item Generate the BEV segmentation maps $\mathbf{Y}_\text{G1}$ and $\mathbf{Y}_\text{G2}$ based on the messages from the two groups, respectively. 
    \item Calculate the CCLoss between the ego CAV and two groups, respectively, and verify if there are malicious CAVs in the two groups.
    \item If the group has no malicious CAVs, mark all CAVs in the group as benign. Otherwise, repeat 2-5 until all malicious CAVs are found or enough benign CAVs are obtained.
\end{enumerate} 
The pseudo-code of PASAC is shown in Algorithm \ref{alg:pasac}.

\begin{table*}[t]
    \caption{\textbf{Quantitative results of CP-Guard} with CCLoss threshold $\varepsilon = 0.08$, PGD iterations = 15, C\&W iteration = 15, and  FGSM noise variance = 10.}
    \label{tab:quantitative_results}
    \renewcommand\arraystretch{1.2}
    \resizebox{1\linewidth}{!}{
    \begin{tabular}{p{5cm}|ccccccc|c}
        \hline 

        { \textbf{Method} }& {Vehicle} & {Sidewalk} & {Terrain} & {Road} & {Buildings} & {Pedestrian} & {Vegetation} & \textbf{mIoU} \\
        \hline \hline
        Upper-bound & 55.58 & 48.20 & 47.33 & 69.60 & 29.34 & 21.67 & 41.02 & 40.45 \\
        
        CP-Guard (against FGSM attack) & 52.76 & 46.35 & 46.67 & 68.32 & 28.98 & 20.51 & 40.15 & 39.30 \\
        CP-Guard (against C\&W attack) & 49.22 & 44.08 & 44.76 & 65.58 & 30.12 & 20.83 & 39.10 & 37.95 \\
        CP-Guard (against PGD attack) & 52.84 & 46.41 & 46.73 & 68.41 & 29.01 & 20.48 & 40.16 & 39.34 \\
        
        Lower-bound & 47.06 & 42.46 & 43.78 & 64.07 & 30.51 & 21.21 & 37.32 & 37.09 \\\hline
        No Defense (FGSM attack) & 26.80 & 27.21 & 29.05 & 36.41 & 16.44 & 12.05 & 22.99 & 21.57 \\
        No Defense (C\&W attack) & 34.53 & 35.66 & 35.54 & 56.59 & 24.27 & 13.37 & 34.10 & 29.80 \\
        No Defense (PGD attack) & 22.50 & 19.63 & 15.42 & 15.33 & 9.18 & 8.29 & 22.72 & 14.34 \\
        \hline
    \end{tabular}
    }
\end{table*}

\begin{figure*}[t]
    \centering
    \includegraphics[width=1\linewidth]{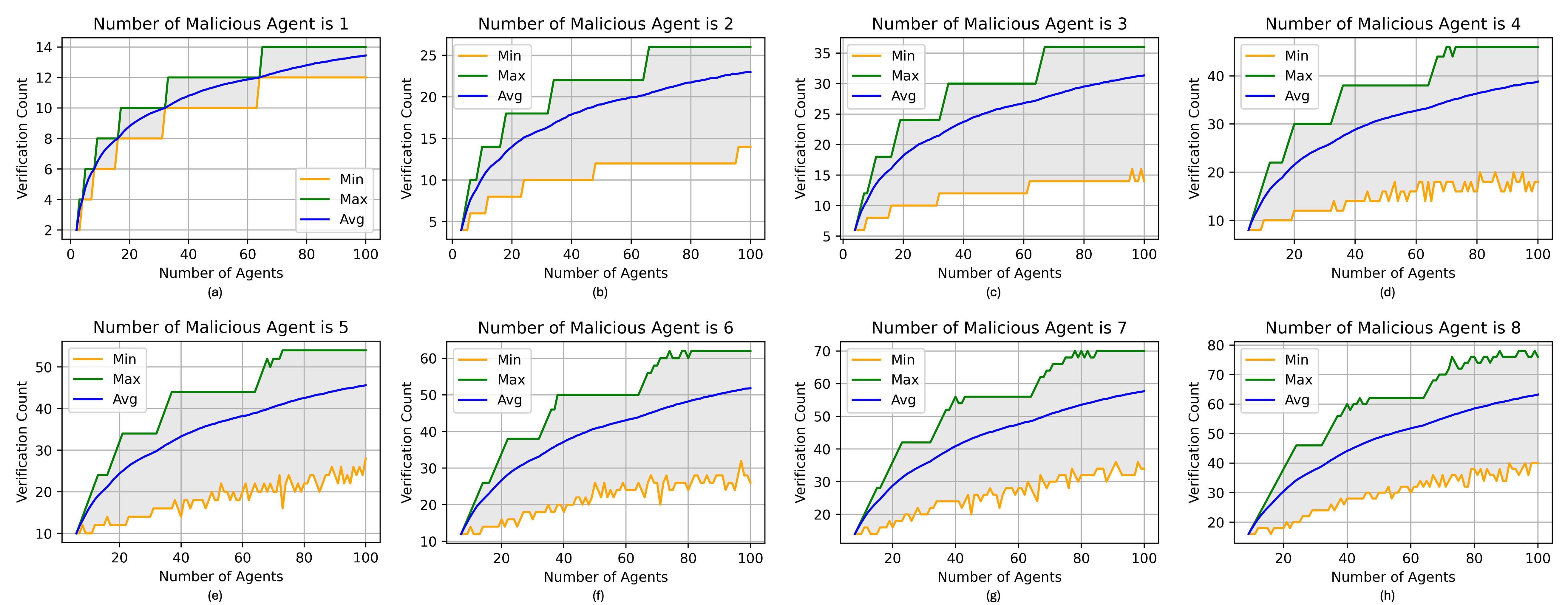}
    \vspace{-5mm}
    \caption{\textbf{Quantitative results of PASAC:} Number of Agents vs Verification Count.}
    \label{fig:pasac_analysis}
    \vspace{-2mm}
\end{figure*}

\subsection{Collaborative Consistency Loss Verification}
\label{subsec:collaborative_consistency_loss_verification}

To verify the consensus between the ego CAV and the collaborators, we design a novel loss function, \textit{Collaborative Consistency Loss} (CCLoss). CCLoss is used to calculate the discrepancy between the ego CAV and the collaborators. 

Given the intermediate feature maps $\mathbf{F}_0$ of the ego CAV and a set of intermediate feature maps $\{\mathbf{F}_1, \ldots, \mathbf{F}_i\}$ from the collaborators. The ego CAV will generate two BEV segmentation maps:
\begin{equation}
    \mathbf{Y}_0=f_\text{D}(\mathbf{F}_0),
\end{equation}
\begin{equation}
    \mathbf{Y}_\text{fuse}=f_\text{D}(f_\text{A}(\mathbf{F}_0, \mathbf{F}_1, \ldots, \mathbf{F}_i)),
\end{equation}
where $\mathbf{Y}_0$ and $\mathbf{Y}_\text{fuse}$ are 3D matrices and their sizes are in $\mathbb{R}^{W_D\times H_D\times C}$ with $W_D$, $H_D$, and $C$ being the width, height, and the number of classes of the BEV segmentation map, respectively. 

As stated before, our key idea is to enable CP to achieve consensus rather than conflict against the ego CAV's perception result. Following this idea, we carefully design the CCLoss to measure the discrepancy between the ego CAV and the collaborators, which is formulated as:
\begin{equation}
    \begin{aligned}
        &\mathcal{L}_\text{CCLoss}(\mathbf{Y}_0, \mathbf{Y}_\text{fuse})=\\
        &\frac{\sum^C_{j=1}w_j\sum^{W_D\cdot H_D}_{i=1}p^0_{i,j}p^\text{fuse}_{i,j}}{\sum^C_{j=1}w_j\left(\sum^{W_D\cdot H_D}_{i=1}p^0_{i,j} + \sum^{W_D\cdot H_D}_{i=1}p^\text{fuse}_{i,j}\right)}
    \end{aligned}
\end{equation}
where $C$ is the number of classes, $p^0_{i,j}$ and $p^\text{fuse}_{i,j}$ are the probabilities of the $j$-th class at the $i$-th pixel in the BEV segmentation map $\mathbf{Y}_0$ and $\mathbf{Y}_\text{fuse}$, respectively, and $w_j$ is the weight of the $j$-th class, defined as the inverse frequency of the class $w_j=1/(\sum^C_{j=1}(p^0_{i,j}+ p^\text{fuse}_{i,j}))^2$. For the numerator of $\mathcal{L}_\text{CCLoss}$, it calculates the weighted sum of the product of the probabilities for each pixel and each class, which essentially measures the overlap between the two distributions. The weight $w_j$ ensures that the contribution of each class is adjusted according to its importance or frequency. The denominator sums up the weighted sums of the probabilities from both the ego CAV's prediction map and the fused segmentation maps for each class. It represents the total probability mass for each class, adjusted by the weights. Finally, the fraction measures the similarity between the two distributions. If these two distributions are similar, the value of $\mathcal{L}_\text{CCLoss}$ will be close to 1. On the contrary, if these two distributions are different, the value $\mathcal{L}_\text{CCLoss}$ will be close to 0.

In addition, we need to set a threshold $\varepsilon$ to determine whether the set contains a malicious agent. Thus, we have the following verification rule if there is a malicious agent in the set:
\begin{equation}
    \mathcal{L}_\text{CCLoss}(\mathbf{Y}_0, \mathbf{Y}_\text{fuse}) \leq \varepsilon.
\end{equation}
The choice of the threshold $\varepsilon$ is crucial. A large $\varepsilon$ will lead to a high false-positive rate, while a small $\varepsilon$ will lead to a high false-negative rate. 


\begin{table*}[t]
    \caption{\textbf{Quantitative results of ablation studies} on $\mathcal{L}_\text{CCLoss}$ threshold $\varepsilon$ against PGD attack (iterations = 15).}
    \label{tab:ablation_results}
    \renewcommand\arraystretch{1.2}
    \setlength{\tabcolsep}{4mm} 
    \resizebox{1\linewidth}{!}{
    \begin{tabular}{c|ccccccc|c}
        \hline 
        \textbf{Threshold} $\varepsilon$ & {Vehicle} & {Sidewalk} & {Terrain} & {Road} & {Buildings} & {Pedestrian} & {Vegetation} & \textbf{mIoU} \\
        \hline \hline
        0.02 & 34.94 & 34.98 & 28.55 & 36.20 & 22.62 & 14.23 & 32.62 & 25.97 \\
        0.05 & 50.70 & 45.25 & 44.76 & 64.50 & 28.34 & 19.84 & 39.45 & 37.79 \\
        0.08 & \textbf{52.84} & \textbf{46.41} & \textbf{46.73} & \textbf{68.41} & \textbf{29.01} & \textbf{20.48} & \textbf{40.16} & \textbf{39.34} \\ 
        0.10 & 52.48 & 46.24 & 46.36 & 67.97 & 28.95 & 20.70 & 39.97 & 39.17 \\
        0.12 & 49.61 & 43.80 & 44.74 & 65.46 & 30.21 & 21.26 & 38.56 & 37.99 \\
        0.15 & 48.62 & 42.89 & 44.19 & 64.91 & 30.25 & 21.19 & 37.75 & 37.52 \\
        \hline
    \end{tabular}
        }
        \vspace{-3mm}
\end{table*}
\begin{table}[t]
    \caption{\textbf{Comparison results} between ROBOSAC and PASAC}
    \label{tab:comparison_results}
    \renewcommand\arraystretch{1.2}
    \resizebox{1\linewidth}{!}{
    \begin{tabular}{c|ccc|ccc}
    \hline
    \multirow{3}{*}{Attack Ratio} & \multicolumn{3}{c|}{ROBOSAC}       & \multicolumn{3}{c}{PASAC (Ours)}        \\ \cline{2-7} 
                                  & \multicolumn{3}{c|}{Verification Count} & \multicolumn{3}{c}{Verification Count} \\
                                  & Min       & Max      & Avg        & Min       & Max      & Avg       \\ \hline\hline
    0.8                           & 1         & 17       & 4.73       & 8         & \textbf{8}        & 8.00       \\ 
    0.6                           & 1         & 46       & 8.29       & 6         & \textbf{8}        & \textbf{7.59}      \\ 
    0.4                           & 1         & 39       & 10.36      & 4         & \textbf{8}        & \textbf{6.60}       \\ 
    0.2                           & 1         & 19       & 4.89       & 4         & \textbf{6}        & \textbf{4.79}      \\ \hline
    Average & 1.00 & 30.25 & 7.06 & 5.50& \textbf{7.50}& \textbf{6.74} \\  \hline

    \end{tabular}
    }
    \vspace{-3mm}
\end{table}

\section{Experiments}

\subsection{Experimental Setup}

\subsubsection{Datasets and Evaluation Metrics.} In our experiments, we leverage V2X-Sim \cite{liV2XSimMultiAgentCollaborative2022} as our dataset. It is the first synthetic dataset for CP generated by CARLA-SUMO co-simulator. In addition, to evaluate the performance of the segmentation, we adopt mean Intersection over Union (mIoU) as the evaluation metric. We also use Verification Count to evaluate the performance of PASAC, which is the total number of times that malicious agents are checked.

\subsubsection{Implementation Details.} We build the collaborative BEV segmentation model by PyTorch, using U-Net \cite{ronnebergerUNetConvolutionalNetworks2015} as the backbone, V2VNet \cite{10.1007/978-3-030-58536-5_36} as the fusion method.
Our experiment is deployed on a computer consisting of 2 Intel(R) Xeon(R) Silver 4410Y CPUs (2.0GHz), four NVIDIA RTX A5000 GPUs, and 512GB DDR4 RAM.
As for the implementation of adversarial attacks, we employ three kinds of attacks: fast gradient sign method (FGSM) \cite{goodfellow2015explainingharnessingadversarialexamples}, Carlini \& Wagner (C\&W) \cite{carlini2017evaluatingrobustnessneuralnetworks}, and the projected gradient descent (PGD) \cite{madry2018towards}. For each attack, 
we set the maximum perturbation $\delta_\text{max} = 0.1$, iterations steps $T = 15$, and the step size $\alpha = 0.01$. 

\subsection{Quantitative Evaluation}
\subsubsection{Evaluation of CP-Guard.} 

We evaluate the efficacy of our CP-Guard scheme against a variety of adversarial attacks. The outcomes of these evaluations are detailed in Table \ref{tab:quantitative_results}. In scenarios where the collaborative perception system lacks defensive mechanism, the mIoU across all three attack modalities significantly falls below the established lower bound, registering at 37.09. This substantial degradation in performance underscores the effectiveness of the adversarial attacks implemented. Conversely, our CP-Guard framework demonstrates robust defensive capabilities, effectively countering all evaluated attacks and achieving an mIoU that closely approaches the upper bound of 40.45.
Specifically, for the FGSM and PGD attacks, setting the CCLoss threshold to $\varepsilon = 0.08$ proves optimal experimentally. This configuration allows CP-Guard to maximally leverage its defensive mechanisms, yielding mIoU scores of 39.30 and 39.34, respectively. The methodology for determining the optimal CCLoss threshold is further explored in the Ablation Studies section of this paper.
In addition, it is noteworthy that while CP-Guard maintains performance above the lower bound when defending against the C\&W attack, the mIoU observed is relatively lower at 37.95. This reduced efficacy can be attributed to the sophisticated fine-grained optimization process inherent to the C\&W attack, which complicates the detection and mitigation efforts.

\subsubsection{Evaluation of PASAC.} To investigate the performance of PASAC, we conduct extensive experiments to study the relationship between the verification count and the number of benign agents and malicious agents. As shown in Fig. \ref{fig:pasac_analysis}, the x-axis represents the number of benign agents and the y-axis represents the verification count. There are three lines in each subfigure, which represent the minimum, average, and maximum verification count, respectively. We can observe that the verification count increases with the number of collaborative agents and the growth trend is fast at the beginning and then becomes slow. In addition, the verification count is far less than the total number of agents, which indicates that PASAC is efficient in sampling collaborators.
For example, when the number of collaborative agents is 100, the number of malicious agents is $m$, the average verification count is usually around  $100\times m\times 0.1$, and the mean verification count is below $100\times m\times 0.05$.

\begin{figure*}[t]
    \centering
    \includegraphics[width=.9\linewidth]{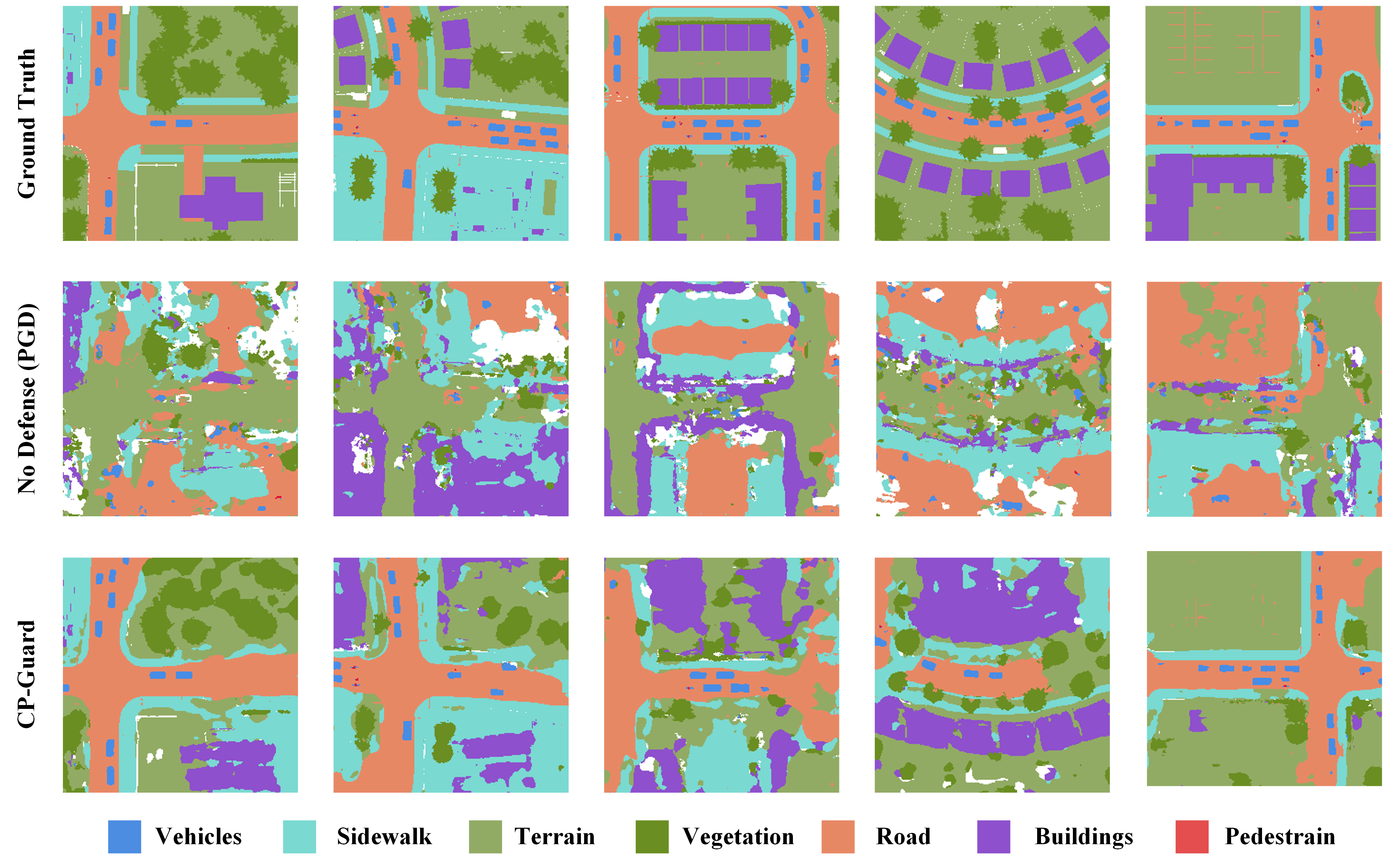}
    \caption{\textbf{Visualization of no defense and defensive CP-Guard results} on V2X-Sim datasets.}
    \label{fig:cp_visualization_3}
    \vspace{-3mm}
\end{figure*}

\subsubsection{Comparison Results.} We conduct a comparative experiment with the previous state-of-the-art method, ROBOSAC \cite{liUsAdversariallyRobust2023}. Following the experiment setup from ROBOSAC, we evaluate the performance of ROBOSAC and PASAC under different attack ratios. The results are shown in Table \ref{tab:comparison_results}. We observe that PASAC outperforms ROBOSAC in terms of the verification count. Specifically, PASAC achieves a lower verification count than ROBOSAC under different attack ratios. For example, when the attack ratio is 0.6, the average verification count of PASAC is 7.59, which is lower than the average verification count of ROBOSAC (8.29), and when the attack ratio is 0.4, the average verification count of PASAC is 6.60, which is much lower than the average verification count of ROBOSAC (10.36).
In addition, the results of PASAC are more stable than ROBOSAC, for example, the maximum verification count of PASAC is 8, while the maximum verification count of ROBOSAC is 46, which is much higher than the average verification count.
These results indicate that PASAC can effectively sample collaborators and outperforms the state-of-the-art method, ROBOSAC. Furthermore, ROBOSAC needs to know the prior probabilities of malicious agents, while PASAC is a probability-agnostic sample method, so it does not require this information, which makes PASAC more practical in real-world scenarios.

\subsection{Qualitative Evaluation}

As depicted in Fig. \ref{fig:cp_visualization_3}, we present the visualization results on the V2X-Sim dataset. Without CP-Guard, attackers can significantly disrupt collaborative perception, leading to a marked degradation in the performance of BEV segmentation tasks.
However, our introduced CP-Guard framework can intelligently identify benign collaborators and eliminate malicious collaborators, thereby facilitating robust CP.

\subsection{Ablation Studies}

We have further undertaken ablation studies to ascertain the optimal CCLoss threshold for our CP-Guard framework in defending against PGD attacks. The results of these studies are summarized in Table \ref{tab:ablation_results}.
On the one hand, when the CCLoss threshold $\varepsilon$ is set below 0.08, the ability of the ego CAV to distinguish malicious agents is progressively impaired, resulting in a decline in the mIoU. Notably, at a CCLoss threshold of $\varepsilon=0.02$, the ego-agent fails to identify the malicious agent, culminating in an mIoU of 25.97, which is substantially below the established lower-bound.
On the other hand, as the CCLoss threshold $\varepsilon$ is increased beyond 0.08, there is a noticeable deterioration in the mIoU, approaching the lower-bound. This suggests that the ego CAV begins to mistrust benign collaborators, increasingly relying on its own inputs. Based on these ablation results, we determine that the optimal CCLoss threshold for CP-Guard against PGD attacks is 0.08. This setting optimally balances the trade-off between excluding malicious inputs and maintaining trust in benign collaborative data, thereby enhancing the overall robustness of the system.

\section{Conclusion}
In this paper, we have designed a novel defense framework for CP named CP-Guard. It consists of two parts, the first is PASAC which can effectively sample the collaborators without the prior probabilities of malicious agents. The second is collaborative consistency loss verification which calculates the discrepancy between the ego CAV and the collaborators, which is used as a verification criterion for consensus. The extensive experiments show that our CP-Guard can defend against different types of attacks and can adaptively adjust the trade-off between the performance and computational overhead.

\section{Acknowledgements}

This work was supported in part by the Hong Kong Innovation and Technology Commission under InnoHK Project CIMDA, in part by the Hong Kong SAR Government under the Global STEM Professorship and Research Talent Hub, and in part by the Hong Kong Jockey Club under the Hong Kong JC STEM Lab of Smart City (Ref.: 2023-0108). The work of Yiqin Deng was supported in part by the National Natural Science Foundation of China under Grant No. 62301300. The work of Xianhao Chen was supported in part by the Research Grants Council of Hong Kong under Grant 27213824.

\bibliography{aaai25, aaai25_2}


\end{document}